%% file: cvpr.tex
\documentclass[final]{cvpr}

\usepackage{times}
\usepackage{epsfig}
\usepackage{graphicx}
\usepackage{amsmath}
\usepackage{amssymb}
\usepackage{autobreak}
\usepackage{multirow}
\usepackage{colortbl}
\usepackage{booktabs}
\usepackage[ruled,vlined]{algorithm2e}
\usepackage{textcomp}
\definecolor{mygray}{gray}{.9}
\definecolor{mygray1}{gray}{.8}

\usepackage[pagebackref=true,breaklinks=true,colorlinks,bookmarks=false]{hyperref}

\def\cvprPaperID{7928} 
\def\confYear{CVPR 2021}
\begin{document}

\title{S2R-DepthNet: Learning a Generalizable Depth-specific Structural Representation}


\author{
{Xiaotian Chen{$^{1}$}\thanks{This work was done when Xiaotian Chen was an intern at Microsoft Research Asia.}} \qquad Yuwang Wang{$^{2}$}\thanks{Corresponding author.} \qquad   Xuejin Chen{$^{1}$} \qquad  Wenjun Zeng{$^{2}$} \qquad\\
\normalsize
$^{1}$\    University of Science and Technology of China ~~ $^{2}$\,Microsoft Research Asia\\
\normalsize
{\tt\small ustcxt@mail.ustc.edu.cn\quad \{yuwwan,wezeng\}@microsoft.com\quad xjchen99@ustc.edu.cn}
}
\maketitle


\maketitle

\begin{abstract}
Human can infer the 3D geometry of a scene from a sketch instead of a realistic image, which indicates that the spatial structure plays a fundamental role in understanding the depth of scenes. 
We are the first to explore the learning of a depth-specific structural representation, which captures the essential feature for depth estimation and ignores irrelevant style information. 
Our S2R-DepthNet (Synthetic to Real DepthNet) can be well generalized to unseen real-world data directly even though it is only trained on synthetic data. 
S2R-DepthNet consists of: a) a Structure Extraction (STE) module which extracts a domain-invariant structural representation from an image by disentangling the image into domain-invariant structure and domain-specific style components, b) a Depth-specific Attention (DSA) module, which learns task-specific knowledge to suppress depth-irrelevant structures for better depth estimation and generalization, and c) a depth prediction module (DP) to predict depth from the depth-specific representation. 
Without access of any real-world images, our method even outperforms the state-of-the-art unsupervised domain adaptation methods which use real-world images of the target domain for training. In addition, when using a small amount of labeled real-world data, we achieve the state-of-the-art performance under the semi-supervised setting. The code
and trained models are available at \url{https://github.com/microsoft/S2R-DepthNet}.

\end{abstract}

\input{intro.tex}

\input{relatedwork.tex}

\input{method.tex}

\input{results.tex}

\section{Conclusion}
In this paper, we present a novel structural representation for generalizable depth estimation. Our learnt representation can be well generalized to unseen real-world images when trained on synthetic data, though there is an obvious style gap. The key is to extract structure information which is disentangled from various domain styles. Since the extracted structure map only contains low-level general structures including a large amount of depth-irrelevant ones, we further propose the DSA module to extract complementary high-level semantic information from image to suppress depth-irrelevant content. The depth-specific structure map works as an information bottleneck and forces the network to infer depth from these essential representations rather than raw images. We even achieve better performance on unseen real-world images than the state-of-the-art domain adaptation methods which uses the real images from target domain for training. As for the limitation, we still need the scenarios (e.g. indoor or outdoor) to be similar between the synthetic and real-world images. This can be addressed by using a general synthetic dataset with various scenarios. 
\vspace{-3mm}
\paragraph{Acknowledgement}
This work was supported in part by the National Natural Science Foundation of China (NSFC) under Grant 61632006 and Grant 62076230; in part by the Fundamental Research Funds for the Central Universities under Grant WK3490000003; and in part by Microsoft Research Asia.

{\small
\bibliographystyle{ieee_fullname}
\bibliography{egbib}
}

\newpage
\appendix
\setcounter{equation}{0}
\setcounter{table}{0}
\setcounter{figure}{0}
\section*{Appendix}
\begin{figure*}[h]
	\centering
	\includegraphics[width=0.88\textwidth]{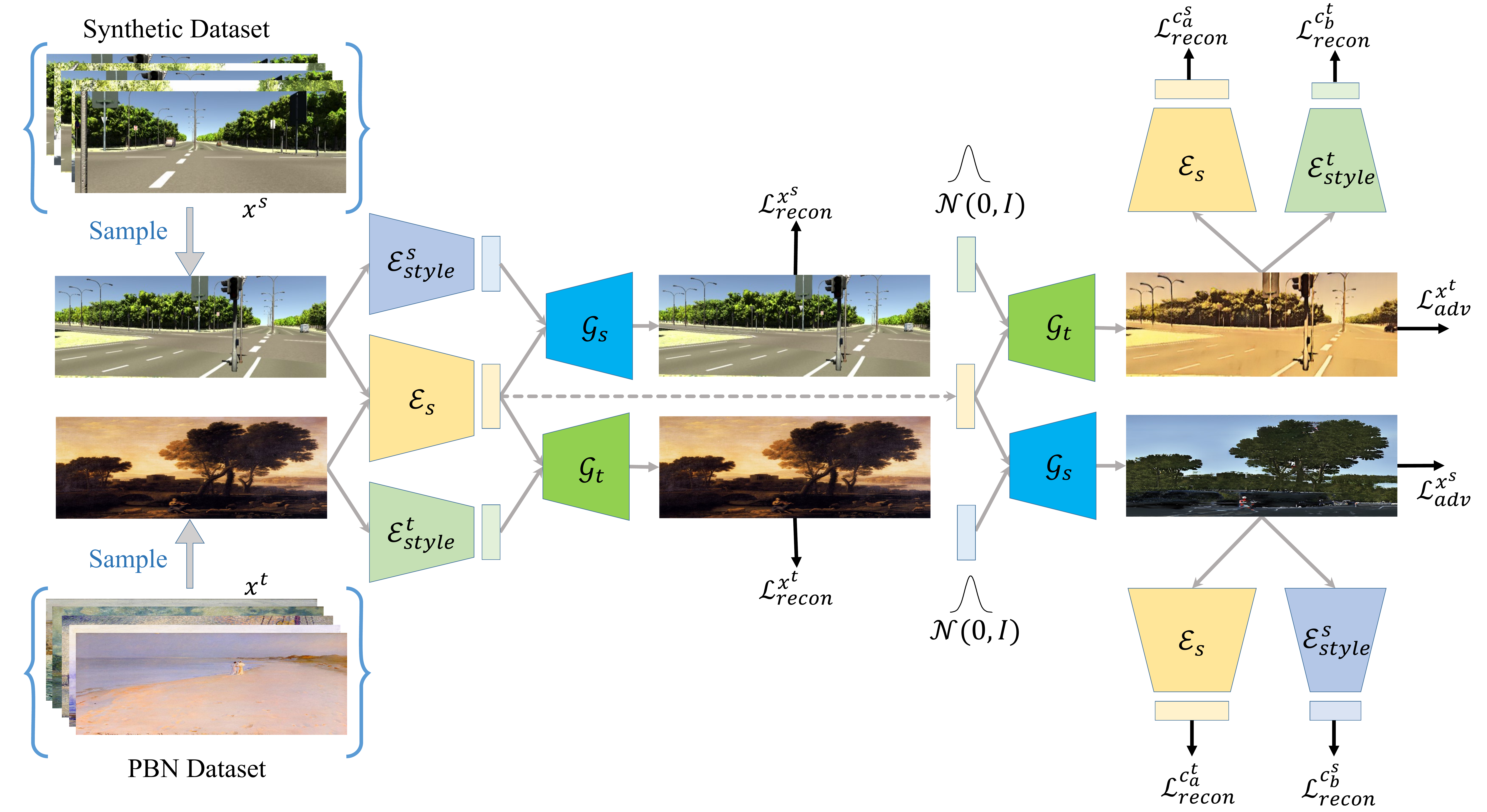}
	\vspace{-1mm}
	\caption{An overview of training process the encoder of STE module. 
	 }
	\label{fig:STEencoder}
	\vspace{0mm}
\end{figure*}

\section{Implementation details of encoder $\mathcal E_s$}
We adopt an image translation framework \cite{huang2018munit} to train the encoder $\mathcal  E_s$ of the STE module,
as shown in Figure \ref{fig:STEencoder}. To make the $\mathcal E_s$ generalizable to various style images, we choose the Painter By Numbers (PBN) dataset \footnote{https://www.kaggle.com/c/painter-by-numbers} with a large style variation as the target domain for image translation and a synthetic dataset as the source domain. 
Given a source domain image $x^s$ and a target domain image $x^t$, 
we first use a shared $\mathcal E_s$ to extract the structure code for both the source and target domains denoted as $c_a^s$ and $c_a^t$ respectively. 
Then the domain specific style encoders $\mathcal E_{style}^s$ and $\mathcal E_{style}^t$ generate style codes $c_b^s$ and $c_b^t$ for the source and target domains respectively. 
The encoded structure code and style code are complementary for each domain. Combining these two codes, the original images from source and target domains can be restored by decoders $G_s$ and $G_t$ respectively. In addition, we also combine $c_a^s$ extracted from the source domain dataset with a randomly sampled style latent code $c_b^t$ from the prior distribution $q(c_b^t) \sim \mathcal N(0,I)$. 
We use $G_t$ to produce the final output image $x_{s\to t}$. 
Similarly, we also combine $c_a^t$ extracted from the target domain dataset with a randomly sampled style latent code $c_b^s$ from the prior distribution $q(c_b^s) \sim \mathcal N(0,I)$. $G_s$ is used to produce the final output image $x_{t\to s}$.

Given an image sampled from the data distribution, because the two parts are complementary, we decode them back to the original image by minimizing
\begin{align}
\begin{autobreak}
\mathcal  L_{recon}^{x^s} = \mathbb E_{x^s \sim p(x^s)}[||\mathcal G_s(\mathcal E_{s}(x^s), \mathcal E_{style}^s(x^s)) - x^s||_1].
\end{autobreak}
\end{align}

After obtaining the final image $x_{s\to t}$ through cross-domain translation, we input it to the shared structure encoder and the specific style encoder, so that the obtained latent code can also be reconstructed by minimizing
\begin{align}
\begin{autobreak}
\mathcal L_{recon}^{c_{a}^{s}} = 
\mathbb E_{c_{a}^{s} \sim p(c_{a}^s), c_{b}^t \sim q(c_{b}^t)}
[||\mathcal E_{s}(\mathcal G_{t}(c_{a}^s, c_{b}^t)) - c_{a}^s||_1],
\end{autobreak}
\end{align}

and

\begin{align}
\begin{autobreak}
\mathcal L_{recon}^{c_{b}^{t}} = 
\mathbb E_{c_{a}^{s} \sim p(c_{a}^s), c_{b}^t \sim q(c_{b}^t)}
[||\mathcal E_{style}^t(\mathcal G_{t}(c_{a}^s, c_{b}^t)) - c_{b}^t||_1],
\end{autobreak}
\end{align}
where  $q({c_{b}^t})$ is the prior $N(0, I)$.
We also use the adversarial loss to match the distribution of the translated images to the PBN data distribution as
\begin{align}
\begin{autobreak}
\mathcal L_{adv}^{x^t} = 
\mathbb E_{c_{a}^{s} \sim p(c_{a}^s), c_{b}^t \sim q(c_{b}^t)}
[log(1- D_t(x_{s\to t}))] + 
\mathbb E_{x_{t}\sim p(x_{t})}[log D_t(x_t)],
\end{autobreak}
\end{align}
where $D_t$ is a discriminator that tries to distinguish translated images from painting images. 
For the other branch, we follow similar pipeline to design the losses.
By minimizing these loss functions, the structure code and style code of the image can be effectively disentangled.
The total training objective is:
\begin{align}
\nonumber \min\limits_{(\mathcal E_s,  \mathcal G_{s}, \mathcal G_{t}, D_{s},  D_{t})} &\max\limits_{(D_{s}, D_{t})}  \mathcal L_{total} = \\
 \nonumber&\mathcal L_{adv}^{x^s} + \mathcal L_{adv}^{x^t} + \lambda_1(L_{recon}^{x^s} + L_{recon}^{x^t})+   \\
&\lambda_2(\mathcal L_{recon}^{c_{b}^{t}} + L_{recon}^{c_{a}^{s}}) + \\
\nonumber &\lambda_3 (\mathcal L_{recon}^{c_{b}^{s}} + L_{recon}^{c_{a}^{t}}),
\end{align}
where $\lambda_1, \lambda_2, \lambda_3$  are weights for the three losses respectively.


\section{Visualize structure maps of images with different styles but the same structure code}
\begin{figure*}[t]
	\centering
	\includegraphics[width=0.88\textwidth]{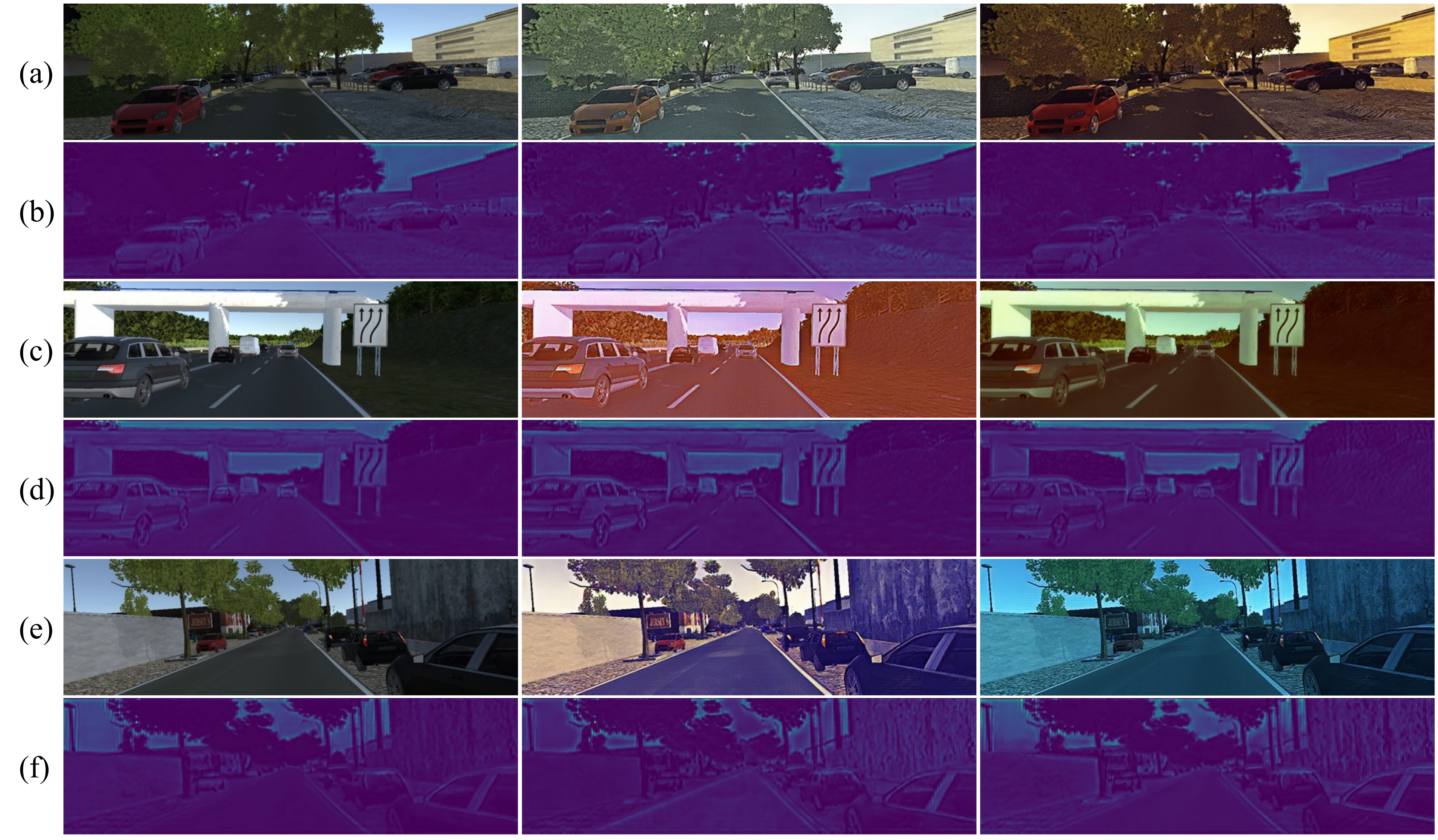}
	\vspace{0mm}
	\caption{Visualize structure maps of different styles and the same structure code. (a), (c) and (e) represent the images of different styles under the same structure code. (b), (d) and (f) are the generated structure maps. 
	 }
	\label{fig:diff_style}
	\vspace{0mm}
\end{figure*}

We show the structure maps of images with different styles but the same structure codes in Figure \ref{fig:diff_style}. The structure maps are generated by feeding images of different styles but same structure code into the STE module.
It can be seen from Figure \ref{fig:diff_style} that although the image styles are different, the generated structure maps are very similar, which further proves that the STE module can ignore the style information and only extract the structure information.

\section{Performance on NYU Depth v2 for semi-supervised setting}
We show the performance of our method under the semi-supervised setting on NYU Depth v2 in Table \ref{tab:nyu-semi}. 
It can be seen from Table \ref{tab:nyu-semi} that even though our method only uses 500 labeled real data (0.42$\%$ of the total dataset) to fine-tune  the domain generalization model, our method outperforms semi-supervised methods \cite{zhao2020domain} under the same settings, and even outperforms some fully supervised methods \cite{li2015depth,eigen2014prediction}. It is worth noting that Laina \etal \cite{laina2016deeper} use more than 120k data to train the depth predictor, and we still surpass their approach on the RMSE.

\section{Datasets and Evaluating Metrics}\label{sec:datasets}
In the following section, we will introduce these datasets and evaluating metrics used in our experiments in detail.

\paragraph{vKITTI \cite{Gaidon2016vkitti}} is a photo-realistic synthetic dataset, that contains 21260 image-depth paired generated from five different virtual worlds in diverse urban settings and weather conditions. The image resolution of this dataset is $375\times1242$. To train our network, 
we follow prior works \cite{zheng2018t2net,zhao2019geometry}, to randomly select 20760 image-depth pairs as our train datasets.  We downsample all the images to $192\times640$ and data augmentation is conducted including random horizontal flipping with a probability of $0.5$, rotation with degrees in $[-5^{\circ}, 5^{\circ}]$, and brightness adjustment.  Because the ground truth of KITTI and vKITTI are significantly different, the maximum depth of the vKITTI dataset is $655.35m$, and the maximum depth value of KITTI is $80m$. In order to reduce the influence of ground truth differences, the depth value of vKITTI is usually clipped to $80m$ \cite{zheng2018t2net,Kundu2018adadepth}.
\paragraph{SUNCG \cite{song2016ssc}} is an indoor synthetic  dataset, which contains 45622 3D houses with various room types.
The image size is $480\times640$. Following previous studies \cite{zheng2018t2net},  we chose the camera locations, poses, and parameters based on the distribution of real NYU Depth v2 dataset \cite{silberman2012indoor} and retained valid depth maps using the same criteria as Zheng \etal \cite{zheng2018t2net}. 130190 image-depth pairs are downsampled to $192\times256$ and used for training.

\input{tbls/nyu_semi}

\paragraph{KITTI \cite{journals/ijrr/GeigerLSU13}} is an outdoor real dataset, which is built for various computer vision tasks for autonomous driving. The images and depth maps are captured for outdoor scenes through a LiDAR sensor deployed on a driving vehicle. The original image resolution is $375\times1241$. 

\paragraph{NYU Depth v2 \cite{silberman2012indoor}} is a real indoor dataset, which contains 464 video sequences of indoor scenes captured with Microsoft Kinect. The dataset is wildly used to evaluate monocular depth estimation tasks for indoor scenes. Following previous work \cite{zheng2018t2net,Kundu2018adadepth,hu2019revisiting,Chen2019structure-aware}, we use the official 654 aligned image-depth pairs for evaluation.
The image resolution is $480\times640$.

\paragraph {Evaluating Metrics} To quantitatively evaluate the proposed approach, we follow previous work \cite{eigen2014prediction,fu2018deep,laina2016deeper,zhou8100183}. The evaluation metrics include root mean squared error~(RMSE), mean relative error (REL), Mean $\log10$ error ($\log10$), root mean squared error in $\log$ space (RMSE$_{\log}$), and squared relative error (Squa-Rel), defined as:
\begin{itemize}
	
	\item RMSE: {\small $\sqrt{\frac{1}{N}\sum_{i=1}^N(d_i - \hat d_i)^2}$}.
	\item REL: {\small $\frac{1}{N}\sum_{i=1}^{N}\frac{|d_i - \hat d_i|}{\hat d_i}$}.
	\item Mean $\log10$ error ($\log10$): {\small $\frac{1}{N}\sum_{i=1}^N|\log_{10}d_i - \log_{10}\hat d_i|$}.
	\item RMSE$_{\log}$: {\small $\sqrt{\frac{1}{N}\sum_{i=1}^N(\log d_i - \log \hat d_i)^2}$}.
	\item Squa-Rel: {\small$\frac{1}{N}\sum_{i=1}^N\frac{|d_i - \hat d_i|^2}{\hat d_i}$}.
	\item Accuracy with threshold $t$: Percentage of pixels whose depth $d_i$ satisfies $\max\big(\frac{d_i}{\hat d_i}, \frac{\hat d_i}{d_i}\big) = \delta < t$, where $t \in [1.25, 1.25^2, 1.25^3]$ respectively.
\end{itemize} 
$d_i$ and $\hat d_i$ are the predicted depth and ground-truth depth at pixel $i$ respectively. $N$ denotes the number of valid pixels in the ground-truth depth map.

\end{document}

%% file: intro.tex
\section{Introduction}\label{sec:introduction}

Monocular depth estimation is a long-standing challenging task, which aims to predict the continuous depth value of each pixel from a single color image. 
This task has a wide range of application in various fields, such as autonomous driving~\cite{Garg2016Unsuper,Godard2017Unsuper}, 3D scene reconstruction~\cite{yao2018mvsnet,Sunghoon2019DPSnet} and robot navigation~\cite{xiazamirhe2018gibsonenv}, etc. Recently, a wide variety of algorithms based on deep convolutional neural networks (DCNNs) have achieved good performance with sufficient amounts of annotated data~\cite{hu2019revisiting, Chen2019structure-aware,xu2017multi,fu2018deep,laina2016deeper,eigen2014prediction, eigen2015predicting}. 
However,  obtaining depth annotations is costly and time-consuming~\cite{Garg2016Unsuper,Godard2017Unsuper,abarghouei18monocular}. Some recent methods have investigated self-supervised depth estimation from stereo images pairs~\cite{Garg2016Unsuper,Godard2017Unsuper} or video sequence ~\cite{zhou8100183,monodepth2,Junsheng2019ICCV} by view reconstruction.  But stereo pairs or video sequences may not always be available in existing datasets. Besides, these models are often limited to the training dataset domain, having difficulty in scaling to various application scenes.


\begin{figure}[t]
	\centering
	\includegraphics[width = \columnwidth]{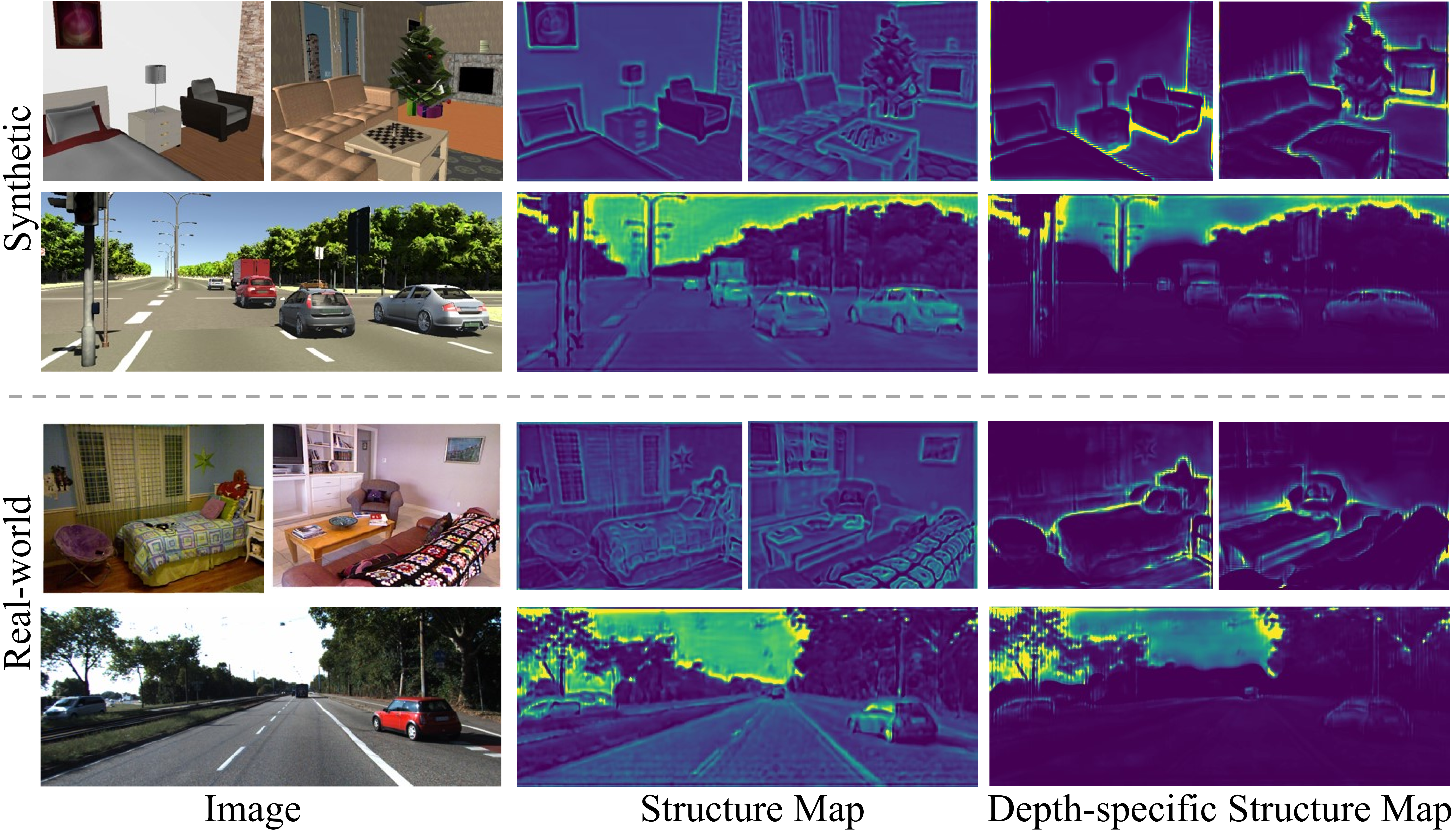}
	
	\caption{
	Visualization of our learnt structural representations.
	It can be seen that even though the input color images from synthetic dataset and real-world dataset are very different in appearance, our structural representations share many similarities, such as layout and object shapes, etc. Furthermore, our depth-specific structure map suppresses the depth-irrelevant structures on the smooth surface, \textsl{e.g.}, lanes on the road and photos on the wall.}
	\label{fig:intro-compare}
	\vspace{-6mm}
\end{figure}

Some researchers switched to use synthetic images~\cite{Gaidon2016vkitti,song2016ssc} for training where depth annotations 
can be acquired directly. 
However, there is usually a domain gap between synthetic data and real-world data, which is caused by style discrepancies across different domains.
To address this issue, some domain adaptation methods~\cite{zheng2018t2net,abarghouei18monocular,Kundu2018adadepth,zhao2019geometry} try to align the feature space of synthetic and real-world images~\cite{zheng2018t2net,Kundu2018adadepth} or translate synthetic images to realistic-looking  ones~\cite{zheng2018t2net,abarghouei18monocular, zhao2019geometry}.
However, these methods all require access to the real-world images of the target domain during the training process, but it is impractical to collect real-world images of various scenes. 

Given the above limitations, we consider a more practical domain generalization scenario. In our setting, we only use a large amount of labeled synthetic data without access of any real-world images of the target domain in the training stage. 
Compared to domain adaptation methods, this is a more difficult task, because we do not even know the style of the real-world images during the training process. 

Aiming for better generalizable depth estimation, we need to seek for the essential representation for this task. Structure information is found to be very important for this task ~\cite{Hu2019VisualizationOC,zhao2019geometry,zhuo2015indoor,Chen2019structure-aware,qi2020geonet++}, and some previous works explore introducing heuristic structure information to network architecture \cite{Chen2019structure-aware,qi2020geonet++} or loss design \cite{zhao2019geometry,Hu2019VisualizationOC,zhuo2015indoor,qi2020geonet++}. We are the first one to explore learning a depth-specific structural representation for generalizable depth estimation. 
The image representation can be decomposed into a domain-invariant structure component and a domain-specific style component~\cite{Kazemi2019WACV,huang2017adain,huang2018munit}. The structure component can be further divided into a depth-specific structure component and a depth-irrelevant structure component. The depth-specific structure component is the most essential for depth estimation and can be effectively transferred from synthetic domain to real-world domain. 

In order to obtain the depth-specific structural representation, we first extract a general domain-invariant structure map from the image using a proposed Structure Extraction (STE) module by decomposing the image into structure and style components inspired by ~\cite{huang2018munit}.
However, the structural representation we thus obtain is a general and low-level image structure, which contains a large amount of depth-irrelevant structures, such as structures on a smooth surface (\textsl{e.g.} lanes on the road or photos on the wall). Furthermore, we propose a Depth-specific Attention (DSA) module to extract high-level semantic information from the input image and help to suppress the depth-irrelevant structures. Since only depth-specific structural information can pass the STE and DSA modules to the depth prediction (DP) module, our S2R-DepthNet trained on synthetic data can be well generalized to unseen real-world images.


We visualise our learnt structural representation and depth-specific structural representation in Figure~\ref{fig:intro-compare}. Even though there is a distinct style difference between the images from synthetic and real-world image dataset, our learnt structure maps and depth-specific structure maps share many similarities. Furthermore, the depth-specific structure map discards depth-irrelevant structures, \textsl{e.g.} lanes. The highlighted sky is an important cue for vanishing point that is helpful for depth estimation, which is similar to~\cite{Hu2019VisualizationOC}.


\textbf{Main contributions:} 
	$(i)$ We are the first to learn a structural representation for generalizable depth estimation, which captures essential structural information and discards style information. S2R-DepthNet can be well generalized to unseen real-world data when only trained on synthetic data.
	$(ii)$ We propose a two-stage structural representation learning pipeline: a general low-level Structure Extraction module to discard style information and a Depth-specific Attention module to suppress depth-irrelevant structure with depth-specific knowledge.
	$(iii)$ We enable a more practical scenario for the depth estimation task, where there is only a large amount of synthetic data but it is hard to acquire real-world data images or depth annotations.

We carry out extensive experiments to demonstrate the effectiveness of our proposed domain-invariant structural representations.  Even though we do not use real-world images for training, our method still outperforms the state-of-the-art domain adaptation methods that use real-world images of the target domain for training.  Surprisingly, when our method uses a small amount of labeled real-world data for training, it also achieves the state-of-the-art performance under the semi-supervised setting.

%% file: relatedwork.tex
\section{Related Work}\label{sec:relatedwork}

\begin{figure*}[t]
	\centering
	\includegraphics[width=0.92\textwidth]{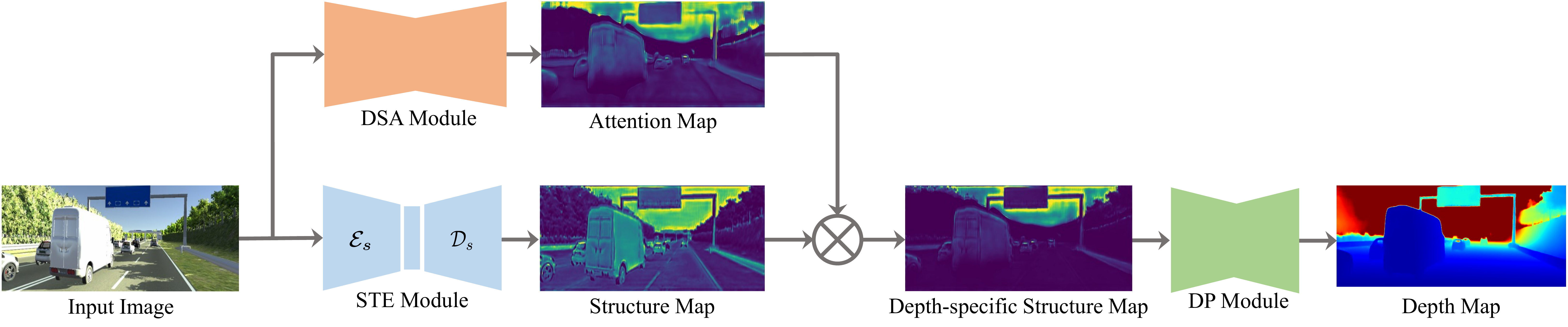}
	\vspace{0mm}
	\caption{ An overview of S2R-DepthNet. Our overall architecture consists of a Structure Extraction (STE) module which extracts a domain-invariant structure map, a Depth-specific Attention (DSA) module which suppresses depth-irrelevant structures by predicting an attention map, and a Depth Prediction (DP) module to predict the final depth map from depth-specific structural representation. $\otimes$ denotes element-wise multiplication.}
	\label{fig:overview}
	\vspace{-5mm}
\end{figure*}

\paragraph{Monocular Depth Estimation.}
The monocular depth estimation task aims to estimate depth from a single image. Previous works explore the network structure~\cite{hu2019revisiting, laina2016deeper,fu2018deep,xu2018structured,Chen2019structure-aware,xia2020generating} or are jointly trained with other tasks, \textsl{e.g.}, normal~\cite{qi2018geonet,yin2019enforcing}, optical flow~\cite{yin2018geonet,chen2019self} and segmentation~\cite{guizilini2019semantically,wang2020sdc}. All these works are trained and tested on datasets of a specific domain without considering domain gaps. It is easy for the method to overfit the specific training dataset and lose the capability of generalization. However, obtaining various depth annotations is costly and time-consuming. Some self-supervised methods~\cite{monodepth2,zhou2019moving,watson2019self,yin2018geonet} take video sequences or stereo pairs as training data and use image warping and reconstruction loss to replace explicit depth supervision. However, dynamic objects and low-texture regions are still a challenge for these methods. Our work is the first one to explore another promising direction, which uses only synthetic data for training and tests directly on real data.
\vspace{-3mm}
\paragraph{Domain Adaptation and Generalization.}
The domain adaptation~\cite{ganin2015unsupervised,long2016unsupervised,saito2018maximum} and generalization~\cite{muandet2013domain, shankar2018generalizing} methods deal with the domain gap between the training and testing datasets. Domain adaptation methods transfer the model trained on the source domain by seeing data in the target domain while domain generalization only performs training on the source domain and tests on the unseen domain. Inspired by the domain adaptation works, researchers try to leverage synthetic data for training a depth estimation model and transfer it to real data. These works can be roughly divided into two categories: aligning the feature space of synthetic and real images~\cite{zheng2018t2net,Kundu2018adadepth,abarghouei18monocular} and translating synthetic images to realistic-looking  ones~\cite{zheng2018t2net,abarghouei18monocular, zhao2019geometry}. Those methods all need access to real images of the target domain, which means it is still necessary to collect real images for training. Our method gets rid of this limitation since we do not use any real images for training and it holds the potential for wider applications.
\vspace{-3mm}
\paragraph{Image Translation and Style Transfer.}
The image translation task translates images from one domain to another~\cite{huang2018munit,gatys2016image,isola2017image}. 
The previous works~\cite{Kazemi2019WACV,huang2017adain, huang2018munit} decompose image representation  into a domain-invariant structure component and a domain-specific style component. Specifically, Kazemi \etal\cite{Kazemi2019WACV} present the style and structure disentangled GAN that learns to disentangle style and structure representations for image generation.  Huang \etal \cite{huang2018munit} disentangle image representation into structure and style codes, and recombine its structure code with a random style code sampled from the style space of the target domain to translate an image to another domain. Targeting for translating images from one style to another, those works extract a structural representation which is disentangled from  domain styles. Our goal is to seek for the essential generalizable structural representation for the depth estimation task by getting rid of the influence of irrelevant factors.

%% file: method.tex

\section{Depth-specific Structural Learning}\label{sec:method}
In this section, we first present an overview of our S2R-DepthNet, then introduce its key modules, and finally provide the training procedure.

\subsection{S2R-DepthNet}

Our S2R-DepthNet is a generalizable depth prediction framework based on depth-specific structural representation leaning. As shown in Figure~\ref{fig:overview}, our framework consists of Structure Extraction (STE) module~$\mathcal{S}$, Depth-specific Attention (DSA) module~$\mathcal{A}$ and a Depth Prediction (DP) module~$\mathcal{P}$. 
To capture the essential representation for depth estimation, we design a two-stage learning pipeline: the STE module to extract a domain-invariant structural representation $M_s$ by disentangling the image into structure and style components, and the DSA module to suppress depth-irrelevant structure with task-specific attention $M_a$, resulting in a depth-specific structure map $M_{sa}$. Finally, we feed $M_{sa}$ into the DP module~$\mathcal{P}$ to predict the depth.

\begin{figure*}[t]
	\centering
	\includegraphics[width=0.88\textwidth]{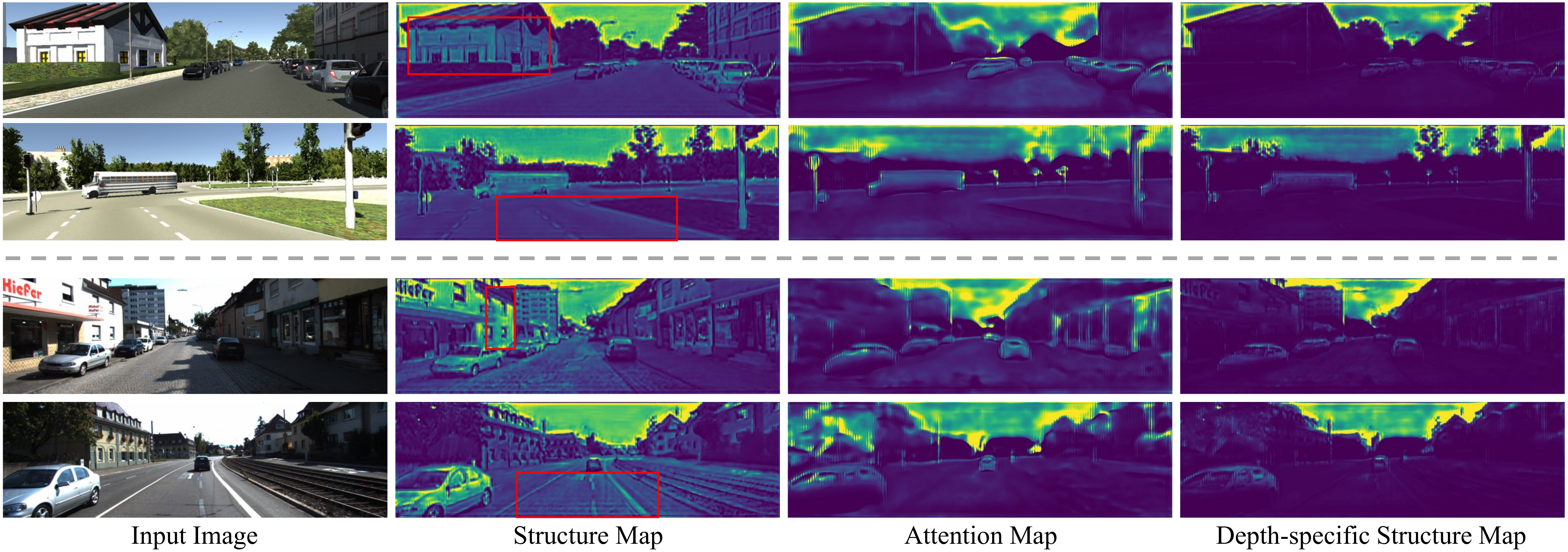}
	\vspace{-1mm}
	\caption{Visualization of intermediate results. The top and bottom two rows are respectively the generated results for synthetic images and real-world images by our S2R-DepthNet trained on synthetic images.
	 }
	\label{fig:structure}
	\vspace{-5mm}
\end{figure*}

\subsection{Structure Extraction Module}

The STE module aims to extract domain-invariant structure information from images with different styles. The STE module consists of an encoder $\mathcal{E}_{s}$ to extract structure information and a decoder $\mathcal{D}_{s}$ to decode the encoded structure information into a structure map as shown in Figure~\ref{fig:overview}.

Inspired by the image translation work~\cite{huang2018munit}, we adopt an image translation framework to train the encoder $\mathcal{E}_{s}$. 
In order to make the $\mathcal{E}_{s}$ generalizable to various style images, we choose the Painter By Numbers (PBN) dataset\footnote{https://www.kaggle.com/c/painter-by-numbers} with a large style variation as the target domain for image translation and the images of a synthetic dataset as the source domain. We use a \textbf{shared} $\mathcal{E}_s$ to extract structure for both the source and target datasets. This is different from~\cite{huang2018munit} which uses different encoders for the source and target datasets. Thus $\mathcal{E}_s$ can see various styles of data and extract structural features that are not sensitive to any specific style. After the training of $\mathcal{E}_s$, we can extract the structure information from the synthetic images. The weights of $\mathcal{E}_s$ are fixed when training other modules to maintain its ability for general structure extraction.

In order to restore the spatial structure of the encoded information of $\mathcal{E}_s$, we choose to use a decoder $\mathcal{D}_s$ to reconstruct a structure map $M_s$ with the same spatial resolution as the input image. Since there is no ground truth of this structure map, we feed the structure map to the DP module, and use the ground truth depth to train $\mathcal{D}_s$. We add a heuristic regularization loss to the structure map by encouraging the value of the structure map to be small wherever the depth map is smooth. Given the input image $I$ and corresponding ground truth $D$, the loss for training $\mathcal{D}_s$ is
\begin{align}

\nonumber\mathcal L_{\mathcal{S}} &= \sum_{p}||\hat D(p) - D(p)||_1 +  \\
&\lambda \sum_{p}||M_s(p)||_1 \cdot e^{-\beta(|\nabla_x D(p) + \nabla_y D(p)|)},\label{equ:train_sd}
\end{align}
where $\hat{D}$ is the predicted depth map ,  $p$ is the index of pixels, $\nabla_x$ and $\nabla_y$ are horizontal and vertical gradient operators respectively, $\lambda $ and  $\beta$ are the hyper parameters.

\subsection{Depth-specific Attention Module}


Our goal is to learn essential structural representation for generalizable depth estimation. Currently the structure map from the STE module contains abundant low level structures including a large amount of depth-irrelevant structures, such as detailed texture structures on smooth surfaces. The encoder of the STE module is designed to capture low-level features with only 4 times downsampling. Besides, there are a lot of Instance Normalization (IN) operations in the encoder for better generalization, which leads to the loss of discriminative  features \cite{huang2017adain,pan2018IBN-Net,Jin_2020_CVPR} and is harmful for semantic feature extraction. In contrast, many researchers have found that high-level semantic knowledge \cite{what3d_cvpr19, Hu2019VisualizationOC,hu2019revisiting,Chen2019structure-aware} is important for the depth estimation task. 

To extract more influential high-level semantic knowledge for depth estimation, we design a DSA module to predict an attention map $A$ from the raw input image. This attention map helps to suppress depth-irrelevant structures by leveraging high-level semantic information extracted from the raw images. We construct the encoder part of the DSA module using the dilated residual network~\cite{Yu2017}, which utilizes dilated convolutions to increase the receptive field while preserving local detailed information. Then we use a decoder to upsample the encoded features to the original resolution and  add a sigmoid layer behind it to generate the attention map. Finally, the obtained attention map is used to weight the general structure map to produce the final depth-specific structure map as:
\begin{align}
M_{sa} = M_s\otimes M_a,
\end{align}
where $\otimes$ denotes element-wise multiplication.

Since we multiply the attention map and the structure map, extra depth-irrelevant information can hardly pass through this bottleneck and the DP module is forced to estimate depth from this concise and comprehensive depth-specific structure representation.
 
We fix the parameters of the previously trained STE module, and train the DSA module to suppress depth-irrelevant structures to get the depth-specific structure map. Meanwhile, the DP module is trained to predict depth from the depth-specific structure map. The loss function for training the DSA and DP modules is:
\begin{align}
\label{equ:train_a}
\mathcal L_{\mathcal{A}} &= \sum_{p}||\hat D(p) - D(p)||_1.
\end{align}

\begin{figure*}[t]
	\centering
	\includegraphics[width=0.88\textwidth]{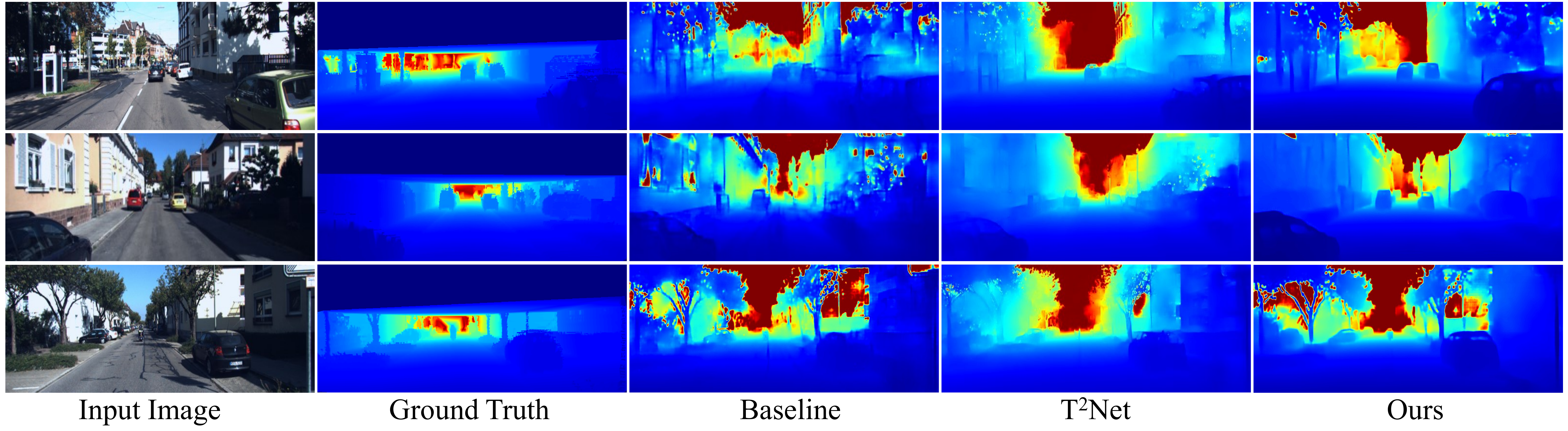}
	\vspace{-1mm}
	\caption{Qualitative comparison of the depth map on the KITTI dataset. The baseline is the DP module only trained on vKITTI dataset. Compared with the approach T$^2$Net\cite{zheng2018t2net} and Baseline, our method can restore clear object boundaries, such as cars and trees.}
	\label{fig:comp-kitti}
    \vspace{-1mm}
\end{figure*}

\subsection{Training Procedure}

In summary, our framework consists of the following modules:  STE module $\mathcal{S}$ which consists of its encoder $\mathcal{E}_s$ and decoder $\mathcal{D}_s$, DSA module $\mathcal{A}$ and DP module $\mathcal{P}$. Instead of training the whole network end-to-end, we design a multi-step training procedure.
1) Train $\mathcal{E}_s$ with the synthetic dataset \cite{Gaidon2016vkitti} and PBN dataset. The detailed network sturcuture and loss are presented in the supplementary. Then Fix $\mathcal{E}_s$ in the following steps.
2) Leave out $\mathcal{A}$, train $\mathcal{D}_s$ and $\mathcal{P}$ with the loss defined in Eq.~\ref{equ:train_sd} on the synthetic dataset. Fix $\mathcal{D}_s$ in the following steps.
3) Involving $\mathcal{A}$, train $\mathcal{A}$ and $\mathcal{P}$ with the loss defined in Eq.~\ref{equ:train_a} on the synthetic dataset.  
After this training process, we get the whole S2R-DepthNet for tesing real-world images.

%% file: results.tex
\section{Experiments}\label{sec:results}

In this section, we first introduce the implementation details and datasets in Section~\ref{sec:details}.
We then conduct experiments on synthetic to real-world generalization task for both the outdoor and indoor scenarios. 
Finally, we provide the ablation studies to analyze the contribution and effectiveness of each module of our framework.



\input{tbls/cmpKITTI}

\subsection{Implementation Details}\label{sec:details}
\paragraph{Network Details.} 
Our S2R-DepthNet consists of three modules:  STE, DSA and DP.
STE module is a standard encoder-decoder architecture. The encoder $\mathcal E_s$ is the same as~\cite{huang2018munit}.
The decoder $\mathcal D_s$ 
includes two up-projection layers ~\cite{laina2016deeper} to restore the encoded structural features to the original image resolution and a convolutional layer to reduce the feature maps into one-channel map.
DSA module is also an encoder-decoder structure where we use a dilated residual network~\cite{Yu2017} as the encoder and three up-projection layers \cite{laina2016deeper} followed by a sigmoid layer as decoder.
For DP module, we follow the depth estimation network architecture of previous works~\cite{zhao2019geometry,zheng2018t2net}.

\vspace{-4mm}
\paragraph{Training Details.} We implement our method based on Pytorch~\cite{paszke2017automatic}. 
We follow the same parameters as~\cite{huang2018munit} to train STE module encoder $\mathcal E_s$. For training the decoder $\mathcal D_s$ of STE module and joint training DSA module and DP module, we use a step learning rate decay policy with Adam optimizer with an initial learning rate of $10^{-4}$. We reduce the learning rate by 50\% every $10$ epochs. We set $\beta_1=0.9, \beta_2=0.999$, weight decay as $10^{-4}$ and the total number of epochs is $60$. The hyper parmameters $\lambda$ and $ \beta$ in Eq.~\ref{equ:train_sd} are set to 1 and 0.001, respectively.
\vspace{-4mm}
\paragraph{Datasets.} For outdoor scenes, we use synthetic Virtual KITTI (vKITTI) ~\cite{Gaidon2016vkitti} as the source domain dataset. 
We use the KITTI dataset \cite{journals/ijrr/GeigerLSU13} as the real-world dataset for evaluation. 
For indoor scenes, we use a synthetic dataset SUNCG as the source domain dataset.
We follow ~\cite{zheng2018t2net} to choose image-depth pairs for training. We use the real-world indoor scene dataset, \textsl{i.e.} NYU Depth v2 ~\cite{silberman2012indoor} for evaluation 
following the same setting with previous works~\cite{zheng2018t2net,Kundu2018adadepth,hu2019revisiting}. 
We follow the previous work ~\cite{eigen2014prediction,fu2018deep,laina2016deeper,zhou8100183} and use standard evaluation metrics.
All results are reported using median scaling as in \cite{Kundu2018adadepth,zhou8100183}, expect that real-world data is used for semi-supervised training.


\input{tbls/cmpKITTI_semi}

\subsection{Experimental results}\label{sec:evaluate results}
Our settings are similar to the typical unsupervised domain adaptation setting, except that we do not access any real-world images. Therefore, we mainly compare with some state-of-the-art domain adaptation methods on depth estimation \cite{Kundu2018adadepth,zheng2018t2net}. We do not compare with~\cite{zhao2019geometry} because it is designed for training with real-world stereo pairs.
We compare two challenging tasks separately: vKITTI \cite{Gaidon2016vkitti} to KITTI \cite{journals/ijrr/GeigerLSU13} and SUNCG \cite{song2016ssc} to NYU Depth v2 \cite{silberman2012indoor}. To better demonstrate the generalizability of our method, we conduct experiments on more auto-driving datasets: Cityscapes~\cite{Cordts2016Cityscapes}, DrivingStereo~\cite{yang2019drivingstereo} and nuSenses~\cite{caesar2020nuscenes}.
In addition, when using a small amount of labeled real-world data, we also compare with state-of-the-art methods under the same semi-supervised setting.


\begin{figure*}[t]
	\centering
	\includegraphics[width=0.9\textwidth]{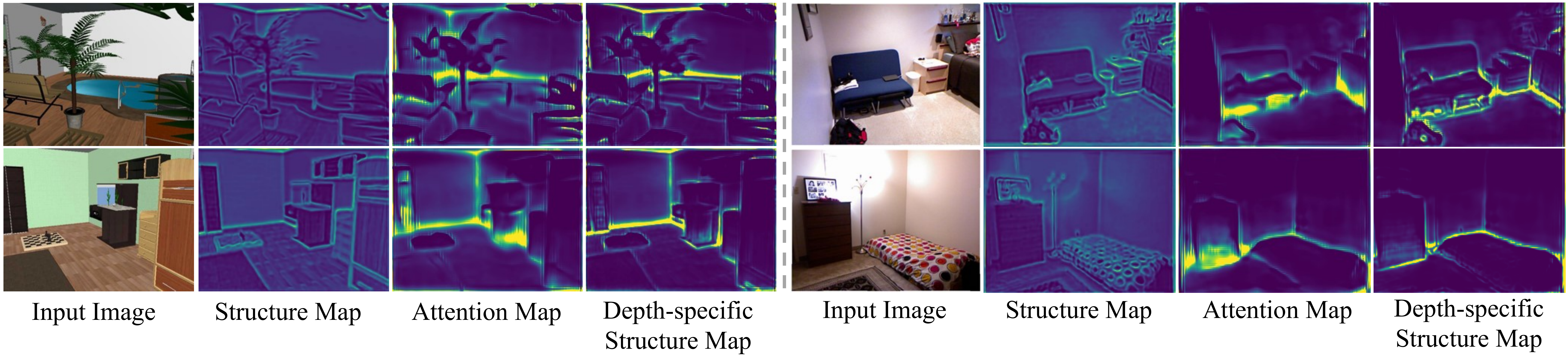}
	\vspace{0mm}
	\caption{Visualization of the structure maps on SUNCG (left) and NYU Depth v2 (right) datasets. Our depth-specific representations focus on the layout, junctions and object boundaries of indoor scenes.}
	\label{fig:nyu_suncg_structure}
	\vspace{-2mm}
\end{figure*}

\input{tbls/cmpNYU}

\begin{table*}[th]
\small
\centering
\caption{
Results on more datasets.}
{%
\begin{tabular}{c||c|c|c|c|c|c|c}
\hline
 &   \multicolumn{3}{c|}{Higher is better} & \multicolumn{4}{c}{Lower is better} \\  \cline{2-8}
\multirow{-2}{*}{Settings}& $\delta < 1.25$ & $\delta < 1.25^2$ & $\delta < 1.25^3$ & Abs Rel & Squa Rel & RMSE & RMSE$_{log}$ \\ \hline

Baseline(vKITTI$\to$Cityscapes)  & 0.516 & 0.747 & 0.854 & 0.297 & 5.077  & 13.938   & 0.452  \\

T$^2$Net(vKITTI$\to$Cityscapes) &0.528&0.760&0.868&0.294&4.639&13.922&0.425\\

Ours(vKITTI$\to$Cityscapes) & \textbf{0.663} & \textbf{0.860} & \textbf{0.941} & \textbf{0.208} & \textbf{2.944} & \textbf{11.164} & \textbf{0.314}  \\
\hline

Baseline(vKITTI$\to$DrivingStereo) & 0.408 & 0.715 & 0.878& 0.374 & 8.619 & 15.822 & 0.440  \\

T$^2$Net(vKITTI$\to$DrivingStereo) & 0.546 & 0.787 & 0.900& 0.302 & 5.689 & 12.892 & 0.377  \\

Ours(vKITTI$\to$DrivingStereo) & \textbf{0.737} & \textbf{0.917} & \textbf{0.971} & \textbf{0.186} & \textbf{2.710} & \textbf{9.166} & \textbf{0.246}  \\
\hline

Baseline(vKITTI$\to$nuScenes) & 0.543 & 0.787 & 0.892 & 0.289 & 3.921 & 11.587 & 0.406  \\

T$^2$Net(vKITTI$\to$nuScenes) & 0.575 & 0.799 & 0.895 & 0.267 & 3.389 & 10.809 & 0.395  \\

Ours(vKITTI$\to$nuScenes) & \textbf{0.601} & \textbf{0.815} & \textbf{0.908} & \textbf{0.249} & \textbf{2.841} & \textbf{10.200} & \textbf{0.366}  \\
\hline

\end{tabular}%
\label{tab:more}
}
\vspace{-4mm}
\end{table*}

\vspace{-3mm}
\paragraph{vKITTI $\rightarrow$ KITTI.} We report experimental results of the proposed method in Table~\ref{tab:kitti}. 
We use the Eigen split~\cite{eigen2014prediction} in the KITTI dataset, which is the same as previous methods \cite{eigen2014prediction, Liu2015TPAMI,zhou8100183,Godard2017Unsuper,Kundu2018adadepth,zheng2018t2net}.
The spatial resolutions are also kept the same.
We take the DP module trained on vKITTI and tested on KITTI as the baseline denoted as DP only (synthetic). We also provide the result of DP module trained and tested on KITTI for reference (denoted as DP only (real-world)), which can be regarded as the upper bound. 
We choose previous state-of-the-art unsupervised domain adaptation methods on depth estimation~\cite{Kundu2018adadepth,zheng2018t2net} that are most similar to our setting for comparison. We also compare to some supervised and self-supervised methods for reference. 
However, it is worth noting that even though our method does not use any real-world images for training, it still outperforms the current state-of-the-art unsupervised domain adaptation methods on the depth estimation task. Results from T$^2$Net~\cite{zheng2018t2net} are recomputed using median scaling with the official pretrained model for fair comparison.
Specifically, compared with T$^2$Net~\cite{zheng2018t2net}, our method improves on $\delta < 1.25$ by $3.17\%$ at cap of 80m and $2.59\%$ at cap of 50m. The Abs-Rel is reduced by $3.51\%$ at cap of 80m and $3.66\%$ at cap of 50m. RMSE is reduced by $4.19\%$ at cap of 80m and $3.31\%$ at cap of 50m.  Even though we do not see the style of real-world images of testing dataset, our proposed method still outperforms the state-of-the-art domain adaptation methods that use real-world image for training.  These significant improvements in all the metrics demonstrate that our proposed structural representation has strong generalization capability on unseen scenes.

In order to further verify the effectiveness of our method, we consider another practical scenario in which there is a small amount of real-world ground-truth data available during training~\cite{zhao2020domain,Kundu2018adadepth}. It can be referred to as a semi-supervised setting. For a fair comparison, we follow the previous works~\cite{zhao2020domain,Kundu2018adadepth} and choose the first 1000 ($4.42\%$ of the total dataset) frames in KITTI as the small amount of real-world labeled data used for training. 
The semi-supervised version of our method is fine-tuned  with these 1000 frames of labeled real-world data based on the domain generalization model. The quantitative results are reported in Table~\ref{tab:kitti_semi}. Our method achieves the best performance on  all the metrics at both 80m and 50m caps compared with previous methods under the same semi-supervised setting~\cite{zhao2020domain,Kundu2018adadepth}. Specifically, compared with Zhao~\etal \cite{zhao2020domain}, our method reduces Abs-Rel by $18.9\%$, RMSE by $6.1\%$ and improves $\delta < 1.25$ by $7.79\%$ at cap of 80m. When compared with Kundu \etal \cite{Kundu2018adadepth}, the Abs-Rel is decreased by $31.5\%$,  RMSE is decreased by $20.3\%$ and $\delta < 1.25$ is increased by $11.0\%$ at cap of 50m.
It is worth noting that even though we do not use any real-world data, our domain generalization (DG) version still outperforms Kundu \etal \cite{Kundu2018adadepth}'s semi-supervised version at cap of 80m and 50m. Kuznietsov \etal \cite{Kuznietsov2017Semi} use the 7346 ($32.5\%$ of total dataset) image-depth pairs and 12600 stereo pairs for training.
We still achieve comparable performance with much less real-world data. 

As Figure \ref{fig:structure} shows, it is obvious that the learned structure maps preserve the fine scene structures. However, these structures contain a lot of depth-irrelevant structures highlighted with red boxes, such as lane lines, textures on houses, etc. The attention maps focus on the object region with geometric structures (such as cars), layout (house silhouette and road guardrail, etc.). In the depth-specific structural maps, many depth-irrelevant structures (lanes on the road and logos on the sign) are suppressed. Another interesting observation is the stronger response in the sky region. The sky with farthest depth value indicates vanishing point, which is an important cue for depth estimation and can be regarded as a kind of strong depth-specific information. 


We also provide qualitative comparisons in Figure \ref{fig:comp-kitti}. Our predicted depth map can restore clear object boundaries, such as cars, trees, and even the structure of tiny objects, which further demonstrates that our structural representations carry essential information for depth prediction.

\vspace{-3mm}
\paragraph{SUNCG $\rightarrow$ NYU Depth v2.} Compared with outdoor scenes, indoor scenes have more various spatial structures and more diverse object categories. 
We compare with a depth estimation method based on domain adaptation~\cite{zheng2018t2net}. We also list some deep learning based supervised depth estimation methods~\cite{eigen2014prediction,li2015depth} for reference, which use 120K real-world image-depth pairs to train the models. As Table~\ref{tab:comnyu} shows, our method significantly outperforms the domain adaptation method \cite{zheng2018t2net} which uses real images of NYU Depth v2 for training. 
\begin{figure}[t]
	\centering
	\includegraphics[width=0.45\textwidth]{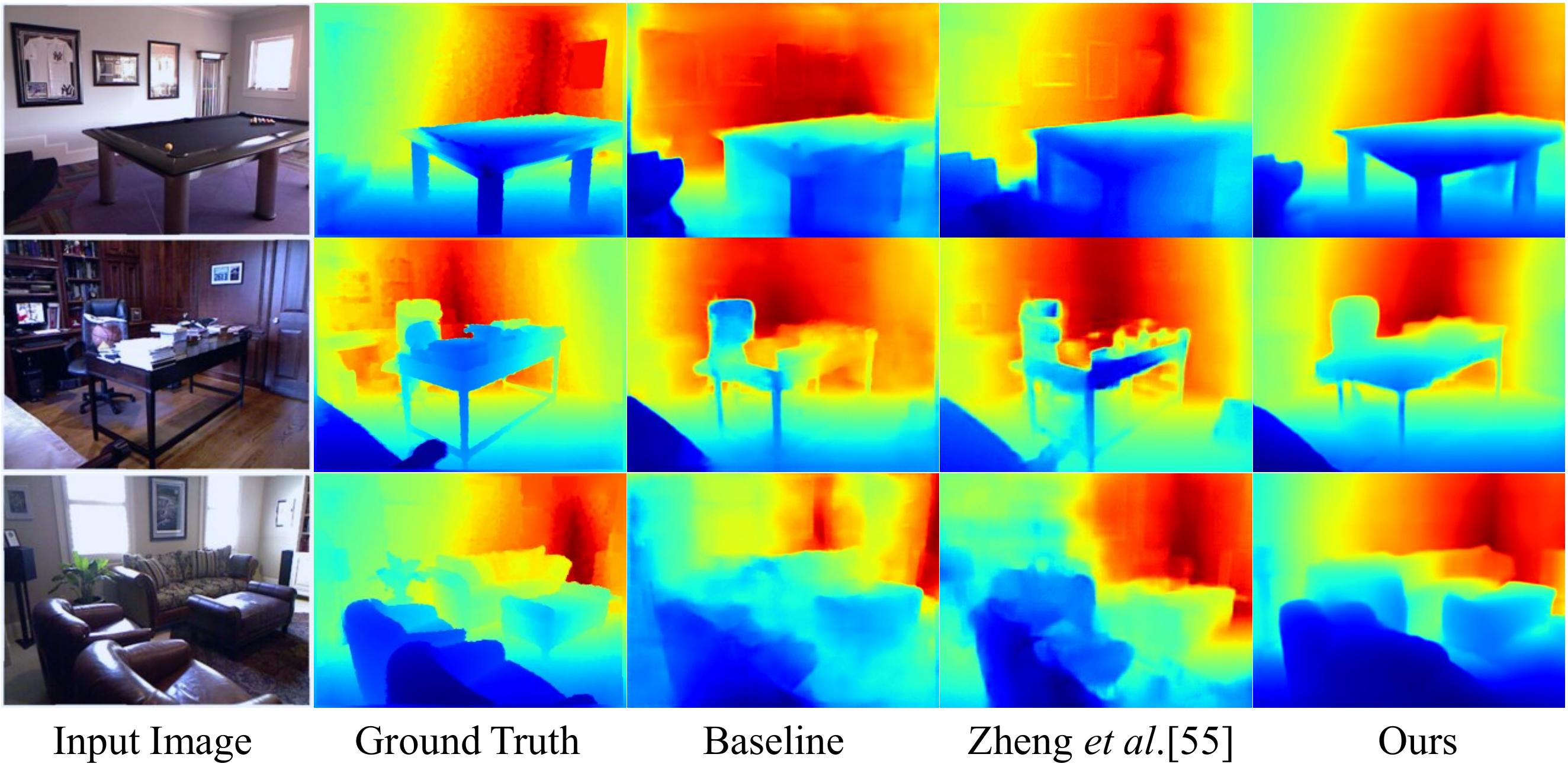}
	\caption{ Qualitative comparison of the depth maps on the NYU Depth v2 dataset.
	 Our method restores clear object boundaries, e.g., sofas and tables.}
	 \vspace{-1mm}
	\label{fig:comp-nyu}
	\vspace{-3mm}
\end{figure}

\input{tbls/abl}

\begin{figure}[h]
	\centering
	\includegraphics[width=0.45\textwidth]{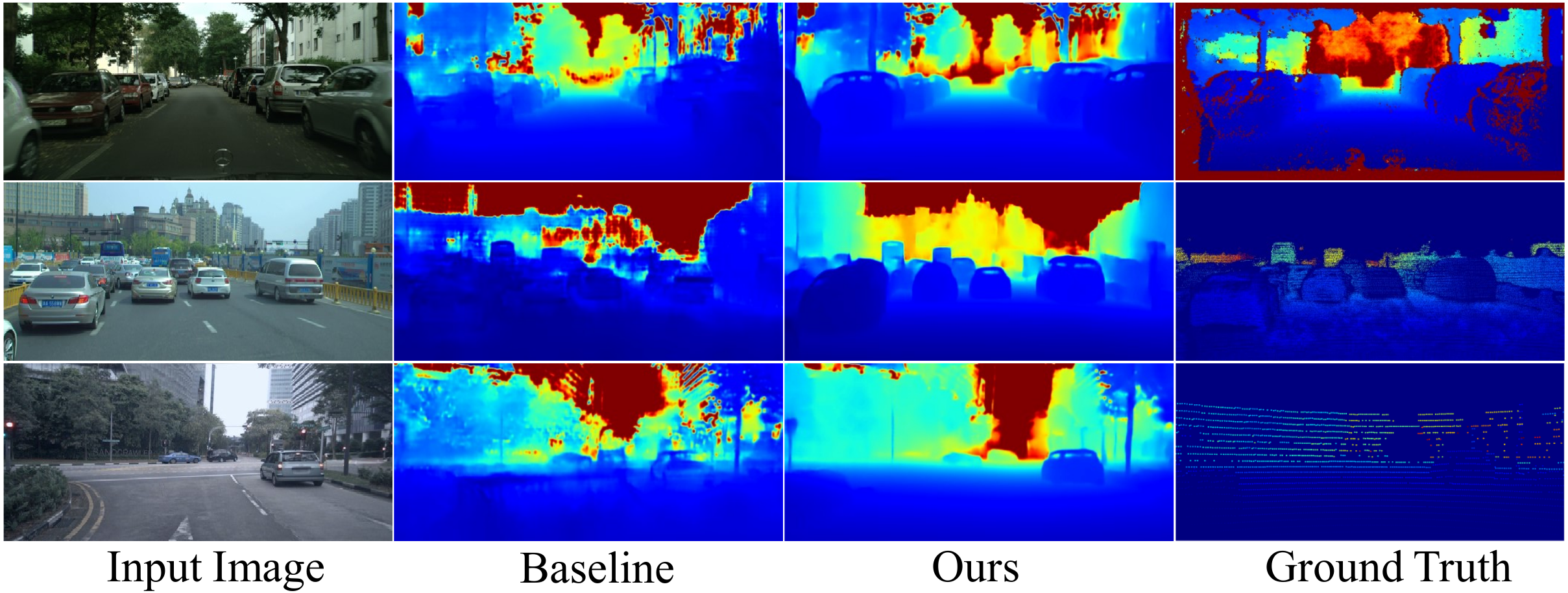}
	\vspace{-1mm}
	\caption{Qualitative results on the cross-datasets. From top to bottom: Cityscapes/DrivingStereo/nuScenes.}
	 \vspace{-2mm}
	\label{fig:comp-visual}
	\vspace{-2.8mm}
\end{figure}

We also visualize the representations in Figure \ref{fig:nyu_suncg_structure}. Interestingly, our depth-specific representations focus on the layout, junctions and object boundaries of indoor scenes.  It is worth noting that the boundaries here are not edge maps but some important boundaries that can clearly reflect the geometric structure of the object. Compared with the structure map, the depth-specific structure map suppresses a large number of structures that are not related to depth information, such as photos on the wall and the texture structure of the floor, etc. Similar characteristics are observed for NYU Depth v2. This also validates that the structural representations reflect the most essential information of depth estimation, which can be effectively transfered between various domains. 
 In Figure \ref{fig:comp-nyu}, we provide qualitative comparisons showing that our results are visually better that other methods. For example, the tables in the first and third rows and the sofas in the second and fourth rows maintain the finer geometric structure and object boundaries.

\vspace{-3mm}

\paragraph{Generalization Results on More Datasets}

Here we further verify our method on the cross-dataset tasks, i.e., vKITTI $\rightarrow$ Cityscapes/DrivingStereo/nuScenes. These three target datasets are auto-driving scenes, but the domain gap is larger compared to vKITTI $\rightarrow$ KITTI due to the variety of camera, location, whether and so on. We follow the same setting as vKITTI $\rightarrow$ KITTI.
As shown in Table~\ref{tab:more} and Figure~\ref{fig:comp-visual}, even though the domain gap is larger, our method consistently achieves significantly better generalizabilty than the Baseline and UDA method T$^2$Net. The improvements vary depending on the structural domain gap.

\subsection{Ablation Study}\label{sec:ablation study}


 For the outdoor datasets, we use the DP module with raw image input as our baseline. We present the results of vKITTI$\rightarrow$KITTI in Table~\ref{tab:abl}. The performance is gradually improved by incorporating STE module and DSA module.
More specifically,  after adding STE module, all the metrics are improved by a large margin from the baseline, where $\delta < 1.25$ is increased by $13.7\%$, Abs-Rel is reduced by $14.8\%$ and RMSE is reduced by $6.47\%$, which shows that removing the style information in the original images can effectively improve the generalization ability of the depth estimation task.
After adding DSA module, the performance in all metrics has been further improved, $\textsl{i.e.}, $ $\delta<1.25$ is increased by $6.99\%$, Abs-Rel is reduced by $17.9\%$ and RMSE is reduced by $13.8\%$, which shows that the removal of depth-irrelevant structures can further improve the generalization ability of the model. 

We study two more combinations of structure map and attention map: addtion (+STE + DRA (Add) ) and concatenation (+STE + DRA (Concat)). As Table \ref{tab:abl} shows, our element-wise multiplication model (+STE + DRA ($\otimes$)) is more effective, because we use the predicted attention map to weight the general structure, which can be regarded as a bottleneck, suppressing depth-irrelevant information and enhancing depth-specific information, while the addtion or concatenation operations introduce redundancy and can not act as the bottleneck effectively.

Intuitively, edge map and the semantic segmentation map can also be regarded as structural representations. 
For the edge map, we apply the Sobel operator \cite{Sobel1973}
 to acquire the edge maps corresponding to the vKITTI and KITTI images.
 For the semantic segmentation map, vKITTI dataset provides semantic labels, and we use one of the state-of-the-art semantic segmentation methods \cite{semantic_cvpr19} to predict semantic segmentation maps for KITTI.
As illustrated in Table \ref{tab:abl}, edge maps and semantic segmentation maps can indeed improve the generalization ability of the network. Since our depth-specific representations only contains the most essential information for depth estimation, they have greater advantages over edge maps and semantic segmentation maps 

In addition, we provide ablation study of our method in indoor scenes on SUNCG~\cite{song2016ssc} and NYU Depth v2~\cite{silberman2012indoor} datasets in Table~\ref{tab:comnyu}. We observe similar trends as reported in the outdoor scenario. Specifically, by adding STE module, the performance of our method has been greatly improved. 
After adding DSA module, the performance has been further improved. 
It is worth noting that the encoder part of our STE module uses the parameters trained on vKITTI (outdoor scenes) and PBN dataset. Although the indoor and outdoor scenes are very different, STE module can still work well in indoor scenes and learn the corresponding structure map of the indoor scenes.




%% file: tbls/cmpKITTI.tex
\begin{table*}[th]
\centering
\caption{\textbf{Performance on KITTI.} All results on KITTI dataset use the Eigen split \cite{eigen2014prediction}. K represents KITTI dataset, V is vKITTI dataset, cap means different gt/predicted depth range. For the supervision or not, Yes represents supervised learning, SSL: self-supervised learning, DG: Domain generalization and UDA: the unsupervised domain adaptation. The best results on each metric are marked in bold.}
\resizebox{\textwidth}{!}{%
\begin{tabular}{c||c|c|c||c|c|c|c|c|c|c}
\hline
 &  &  & & \multicolumn{3}{c|}{Higher is better} & \multicolumn{4}{c}{Lower is better} \\  \cline{5-11}
\multirow{-2}{*}{Method} & \multirow{-2}{*}{Dataset} & \multirow{-2}{*}{Supervision} & \multirow{-2}{*}{cap} & $\delta < 1.25$ & $\delta < 1.25^2$ & $\delta < 1.25^3$ & Abs Rel & Squa Rel & RMSE & RMSE$_{log}$ \\ \hline
Eigen \etal \cite{eigen2014prediction} & K & Yes & 0$\sim$80m & 0.692 & 0.899 & 0.967 & 0.215 & 1.515 & 7.156 & 0.270 \\ 


Zhou \etal \cite{zhou8100183} & K (Video) &SSL & 0$\sim$80m & 0.678 &0.885&0.957&0.208 &1.768 & 6.856 &0.283 \\ 

Godard \etal \cite{Godard2017Unsuper}  & K (Stereo)& SSL & 0$\sim$80m & 0.803 & 0.922 & 0.964 & 0.148 & 1.344 & 5.927 & 0.247\\  \hline

\rowcolor{mygray}
DP only (synthetic) &V & DG &0$\sim$80m &0.642 & 0.861 & 0.944 &0.236&2.171&7.063&0.315 \\ 
\rowcolor{mygray}
DP only (real-world) & K & Yes &0$\sim$80m& 0.804&0.935&0.977&0.141&0.980&5.224&0.217 \\ 
\rowcolor{mygray}
Kundu \etal \cite{Kundu2018adadepth} & V & UDA &0$\sim$80m&  0.665&0.882&0.950&0.214&1.932&7.157&0.295 \\ 
\rowcolor{mygray}
T$^2$Net \cite{zheng2018t2net} & V & UDA & 0$\sim$80m& 0.757 & 0.918 & 0.969 & 0.171 & 1.351 & 5.944 & 0.247 \\
\rowcolor{mygray}
Ours &V& DG & 0$\sim$80m& \textbf{0.781} & \textbf{0.931} & \textbf{0.972} & \textbf{0.165} & \textbf{1.351} & \textbf{5.695} & \textbf{0.236} \\ \hline
\hline

Zhou \etal \cite{zhou8100183} & K  (Video)& SSL& 0$\sim$50m & 0.735 & 0.915 & 0.968 & 0.190 & 1.436 & 4.975 & 0.258 \\ 
Godard \etal \cite{Godard2017Unsuper}  & K  (Stereo)& SSL& 0$\sim$50m & 0.818 & 0.931 & 0.969 & 0.140 & 0.976 & 4.471 & 0.232\\ \hline
\rowcolor{mygray}
DP only (synthetic) & V& DG&0$\sim$50m &0.654&0.872&0.950&0.229&1.726&5.539&0.301   \\ 
\rowcolor{mygray}
DP only (real-world) & K & Yes &0$\sim$50m &  0.819 & 0.944 & 0.980 & 0.136&0.787 & 3.978 &0.205  \\ 
\rowcolor{mygray}
Kundu \etal \cite{Kundu2018adadepth} &V& UDA & 0$\sim$50m& 0.687 & 0.899 & 0.958 & 0.203 & 1.734 & 6.251 & 0.284 \\ 
\rowcolor{mygray}
T$^2$Net \cite{zheng2018t2net} & V & UDA & 0$\sim$50m& 0.773 & 0.928 & 0.974 & 0.164 & 1.019 & 4.469 & 0.231 \\ 
\rowcolor{mygray}
Ours &V& DG & 0$\sim$50m& \textbf{0.793 }& \textbf{0.939 } & \textbf{0.976 }& \textbf{0.158 }& \textbf{1.000 }& \textbf{4.321 } & \textbf{0.223 }\\ \hline

\end{tabular}%
\label{tab:kitti}
}
\vspace{-5mm}
\end{table*}

%% file: tbls/cmpKITTI_semi.tex
\begin{table*}[th]
\small
\centering
\caption{\textbf{Performance on KITTI for semi-supervised setting.} All results on KITTI dataset use the Eigen split \cite{eigen2014prediction}. K represents KITTI dataset,V is vKITTI dataset, cap means different gt/predicted depth range. The best results on each metric are marked in bold.}
\resizebox{\textwidth}{!}{%
\begin{tabular}{c||c|c||c|c|c|c|c|c|c}
\hline
 &  &  &  \multicolumn{3}{c|}{Higher is better} & \multicolumn{4}{c}{Lower is better} \\  \cline{4-10}
\multirow{-2}{*}{Method} & \multirow{-2}{*}{Dataset} & \multirow{-2}{*}{cap} & $\delta < 1.25$ & $\delta < 1.25^2$ & $\delta < 1.25^3$ & Abs Rel & Squa Rel & RMSE & RMSE$_{log}$ \\ \hline

\rowcolor{mygray}
Kundu \etal \cite{Kundu2018adadepth} &V+K(Small) &0$\sim$80m& 0.771 &0.922&0.971&0.167&1.257&5.578&0.237 \\ 
\rowcolor{mygray}
Zhao \etal \cite{zhao2020domain} &V+K(Small) &0$\sim$80m& 0.796 & 0.922 & 0.968 & 0.143& 0.927& 4.694&0.252 \\ 
\rowcolor{mygray}
Ours-DG &V& 0$\sim$80m& 0.781 & 0.931 & 0.972 & 0.165 & 1.351 & 5.695 & 0.236 \\ 
\rowcolor{mygray}
Ours-S &V+K(Small) &0$\sim$80m& \textbf{0.858 }& \textbf{0.955  } & \textbf{0.984 } & \textbf{0.116 }& \textbf{0.766 }  & \textbf{4.409 } & \textbf{0.185 } \\ \hline
\hline
Kuznietsov \etal \cite{Kuznietsov2017Semi}& K+Stereo & 0$\sim$80m & 0.862 & 0.960 & 0.986 & 0.113 & 0.741 & 4.621 & 0.189 \\ \hline
\hline
\rowcolor{mygray}
Kundu \etal \cite{Kundu2018adadepth} &V+K(Small) &0$\sim$50m& 0.784 &0.930&0.974&0.162&1.041&4.344&0.225 \\ 
\rowcolor{mygray}
Ours-DG &V & 0$\sim$50m&0.793 & 0.939  & 0.976 & 0.158 & 1.000 & 4.321  & 0.223 \\ 
\rowcolor{mygray}
Ours-S &V+K(Small) &0$\sim$50m& \textbf{0.870 }& \textbf{0.959 }& \textbf{0.986 }& \textbf{0.111 } & \textbf{0.642 } & \textbf{3.463 } &\textbf{0.176 } \\  \hline
\hline
Kuznietsov \etal \cite {Kuznietsov2017Semi}& K+Stereo &  0$\sim$50m & 0.861 & 0.964 & 0.989 & 0.117 & 0.597 & 3.531 & 0.183 \\ \hline\hline




\end{tabular}%
\label{tab:kitti_semi}
}
\vspace{-3mm}
\end{table*}

%% file: tbls/cmpNYU.tex
\begin{table}[tb]
\renewcommand\tabcolsep{1.8pt}
	\centering
	\caption{Performance on NYU Depth v2.
}
		\scalebox{0.77}{\begin{tabular}{l|c|c|c|c|c|c}
		\toprule
		Method  & Abs Rel & RMSE & $\log{10}$ & $\delta < 1.25$ & $\delta < 1.25^2$ & $\delta < 1.25^3$ \\
		\midrule
		Li \etal \cite{li2015depth}               & 0.232  & 0.821  & 0.094  & 0.621 & 0.886  & 0.968     \\

		Eigen \etal  \cite{eigen2014prediction}   & 0.215  & 0.907  & -  & 0.611 & 0.887  & 0.971     \\
		\hline		
		\rowcolor{mygray}
		T$^2$Net \cite{zheng2018t2net}  & 0.203 & 0.738 & - & 0.670 & 0.891 & 0.966 \\ 
		\rowcolor{mygray}
		Baseline   & 0.278  & 0.899   & 0.111  & 0.557 & 0.826  & 0.940      \\
		\rowcolor{mygray}
		+ STE & 0.225 & 0.756 & 0.093 & 0.643 & 0.880 & 0.962 \\
		\rowcolor{mygray}
		+ STE + DSA & \textbf{0.196 } & \textbf{0.662 } & \textbf{0.082 } & \textbf{0.695 } &\textbf{0.910 } & \textbf{0.972 }\\
		\bottomrule
	\end{tabular}
	}
	\label{tab:comnyu}
	\vspace{-5mm}
\end{table}

%% file: tbls/abl.tex
\begin{table*}[th]
\small
\centering
\caption{
Comparison of the contribution of each module with best results marked in bold.}
{%
\begin{tabular}{c||c|c|c|c|c|c|c}
\hline
 &   \multicolumn{3}{c|}{Higher is better} & \multicolumn{4}{c}{Lower is better} \\  \cline{2-8}
\multirow{-2}{*}{Method}& $\delta < 1.25$ & $\delta < 1.25^2$ & $\delta < 1.25^3$ & Abs Rel & Squa Rel & RMSE & RMSE$_{log}$ \\ \hline

Baseline  & 0.642 &0.861&0.944&0.236&2.171&7.063&0.315 \\ 

+STE  & 0.730 & 0.903 & 0.958 & 0.201& 1.989& 6.606&0.273 \\ 

+STE + DSA (Add) & 0.751& 0.915 & 0.968 & 0.185 & 1.715 & 6.214 & 0.257 \\

+STE + DSA (Concat) & 0.753& 0.919 & 0.970 & 0.179 & 1.487 & 5.975 & 0.252 \\


+STE + DSA ($\otimes$) & \textbf{0.781 } & \textbf{0.931 }& \textbf{0.972 } & \textbf{0.165 } & \textbf{1.351 } & \textbf{5.695 } & \textbf{0.236 }\\ \hline

Edge + Baseline & 0.688& 0.887 & 0.957 & 0.217 & 1.970 & 6.674 & 0.285 \\ \hline

Segmentation Map + Baseline & 0.689& 0.901 & 0.965 & 0.204 & 1.645 & 6.240 & 0.274 \\ \hline




\end{tabular}%
\label{tab:abl}
}
\vspace{-5mm}
\end{table*}

%% file: tbls/nyu_semi.tex
\begin{table}[tb]
\renewcommand\tabcolsep{1.8pt}
	\centering
	\caption{Performance on NYU Depth v2 for semi-supervised setting with best results  marked in bold.
}
		\scalebox{0.77}{\begin{tabular}{l|c|c|c|c|c|c}
		\toprule
		Method  & Abs Rel & RMSE & $\log{10}$ & $\delta < 1.25$ & $\delta < 1.25^2$ & $\delta < 1.25^3$ \\
		\midrule
		Li \etal \cite{li2015depth}               & 0.232  & 0.821  & 0.094  & 0.621 & 0.886  & 0.968     \\
		\hline

		Eigen \etal  \cite{eigen2014prediction}   & 0.215  & 0.907  & -  & 0.611 & 0.887  & 0.971     \\
		\hline
		Laina \etal \cite{laina2016deeper} & 0.127& 0.573& 0.055 & 0.811&0.953&0.988 \\
		\hline
		\rowcolor{mygray}
		Zhao \etal \cite{zhao2020domain} & 0.186 & 0.710 & - & 0.712 & 0.917 & 0.977 \\
		\rowcolor{mygray}
		 Ours & \textbf{0.168 } & \textbf{0.544 } & \textbf{0.069 } & \textbf{0.764 } &\textbf{0.945 } & \textbf{0.984 }\\
		\bottomrule
	\end{tabular}
	}
	\label{tab:nyu-semi}
	\vspace{-5mm}
\end{table}